\documentclass{article}

\PassOptionsToPackage{numbers, compress}{natbib}

\usepackage[preprint]{neurips_2025}

\usepackage[utf8]{inputenc} 
\usepackage[T1]{fontenc}    
\usepackage{hyperref}       
\usepackage{url}            
\usepackage{booktabs}       
\usepackage{amsfonts}       
\usepackage{nicefrac}       
\usepackage{microtype}      
\usepackage{xcolor}         

\usepackage{wrapfig}
\usepackage{graphicx}
\usepackage{amsmath}
\usepackage{multirow}
\usepackage{adjustbox}
\usepackage{xcolor,colortbl}
\usepackage{tabularx}
\usepackage{subcaption}
\usepackage{amsthm}
\usepackage{authblk}

\DeclareUnicodeCharacter{200B}{}
\newtheorem{lemma}{Lemma}
\newtheorem{theorem}{Theorem}
\definecolor{lightgreen}{RGB}{40, 150, 40}  
\definecolor{lightblue}{rgb}{0.25, 0.6, 1.0}
\definecolor{lightred}{rgb}{1.0, 0.4, 0.4} 

\def\eg{\emph{e.g.}}

\title{CAD: A General Multimodal Framework for Video Deepfake Detection via Cross-Modal Alignment and Distillation}

\author{%
  \textbf{Yuxuan Du}\textsuperscript{1*},
  \textbf{Zhendong Wang}\textsuperscript{2*},
  \textbf{Yuhao Luo}\textsuperscript{3*},
  \textbf{Caiyong Piao}\textsuperscript{3*}, \\
  \textbf{Zhiyuan Yan}\textsuperscript{1},
  \textbf{Hao Li}\textsuperscript{1},
  \textbf{Li Yuan}\textsuperscript{1$^\dagger$}
}

\affil{
  {\tt 
  $*$ Equal Contributors, $\dagger$ Corresponding Authors
  }
  \par
  \textsuperscript{1} Peking University, Shenzhen Graduate School, \\
  \textsuperscript{2} University of Science and Technology of China, \\
  \textsuperscript{3} Chinese University of Hong Kong, Shenzhen
  \\
  {\tt 
  dyx2290246176@gmail.com, yuanli-ece@pku.edu.cn
  }
}

\begin{document}

\maketitle
\begin{abstract}
The rapid emergence of multimodal deepfakes~(visual and auditory content are manipulated in concert) undermines the reliability of existing detectors that rely solely on modality‐specific artifacts or cross‐modal inconsistencies.
In this work, we first demonstrate that modality‐specific forensic traces (e.g., face‐swap artifacts or spectral distortions) and modality‐shared semantic misalignments (e.g., lip‐speech asynchrony) offer complementary evidence, and that neglecting either aspect limits detection performance.
Existing approaches either naively fuse modality-specific features without reconciling their conflicting characteristics or focus predominantly on semantic misalignment at the expense of modality-specific fine‐grained artifact cues.
To address these shortcomings, we propose a general multimodal framework for video
deepfake detection via \textbf{C}ross-Modal \textbf{A}lignment and \textbf{D}istillation~(CAD). 
CAD comprises two core components: 1) Cross-modal alignment that identifies inconsistencies in high-level semantic synchronization (\eg, lip-speech mismatches); 2) Cross-modal distillation that mitigates feature conflicts during fusion while preserving modality-specific forensic traces (\eg, spectral distortions in synthetic audio). Extensive experiments on both multimodal and unimodal (\eg, image-only/video-only)deepfake benchmarks demonstrate that CAD significantly outperforms previous methods, validating the necessity of harmonious integration of multimodal complementary information.
\end{abstract}

\section{Introduction}
The rapid advancement of deepfake generation has led to highly sophisticated multimodal forgeries that seamlessly integrate manipulated visual, audio, and textual elements to produce realistic and deceptive content~\cite{pei2024deepfake,yan2024df40,seow2022comprehensive}. Modern multimodal generation techniques increasingly exploit cross-modal coherence, such as syncing synthetic voices with lip movements~\cite{prajwal2020wav2lip, zhou2023makeltalk}, generating contextually plausible scripts~\cite{openai2023gpt4,yan2025gpt}, or blending facial reenactments with cloned speech~\cite{Korshunova2017faceswap-cnn, 2021faceswap}, thereby making the \textbf{detection of such multimodal forgeries increasingly challenging} for existing approaches.
Existing unimodal detection methods (\eg, facial blending anomaly detectors~\cite{shiohara2022detecting,cheng2024can,yan2024generalizing} or spectral artifact analysis in audio~\cite{yi2023audio,li2024audio}) mainly rely on isolated artifacts in a single modality, \textit{making them inherently blind to cross-modal discrepancies and failing to identify cross-modal manipulations} like synchronized lip movements with AI-generated voices.


Generally, the multimodal forgeries essentially expose detection traces from two key perspectives. 1) \textbf{\textit{Modality-Specific Inconsistencies:}} These include artifacts confined to a single modality, such as unnatural facial texture blending in manipulated videos (\cite{li2020face,shiohara2022detecting}) or spectral distortions in synthesized audio (\cite{almutairi2022review,muller2022does}). 2) \textbf{\textit{Modality-Shared Semantic Misalignments:}} These involve inconsistencies between modalities, such as mismatches between lip movements and speech (\cite{Haliassos2022Training,Haliassos2021lips}), or incongruent emotional tones across audio and visual streams (\cite{Oorloff2024AVFF}).
However, we find that \textbf{most existing multimodal detectors fail to consider how to fully utilize both modality-specific and shared forgeries for comprehensive detection}.
To illustrate, most multimodal frameworks (such as \cite{liu2024lips, Oorloff2024AVFF, Yang2023AVoiD-DF}) primarily focus on semantic coherence (\eg, lip-speech alignment) but often overlook modality-specific forensic traces, such as unnatural texture blending or flickering in synthesized faces and subtle acoustic artifacts in synthetic speech.
Additionally, approaches like \cite{zhang2024inclusion} directly combine modality-specific cues without effectively leveraging modality-shared information, \textit{missing the opportunity to exploit cross-modal correlations that could strengthen forgery detection.}

\begin{wrapfigure}{r}{0.53\textwidth}
 \vspace{-5mm}
 \centering
 \includegraphics[width=0.5\textwidth, height=5cm]{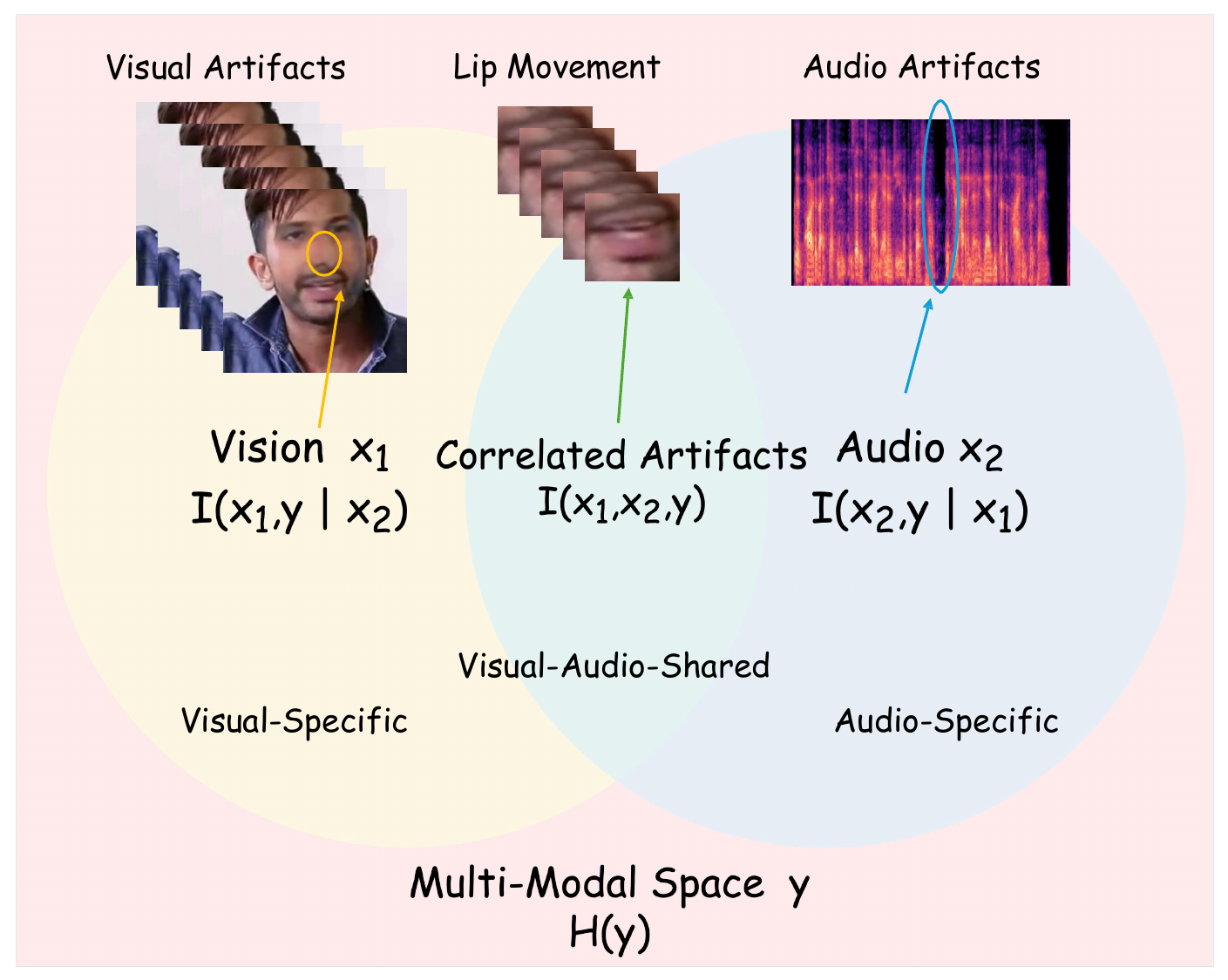}
 \caption{Venn diagram illustrating different types of artifacts in multimodal deepfakes, categorized into \textit{modality-specific} and \textit{modality-shared} cues.
 Specifically, visual artifacts ($x_1$) may include blending boundaries of face-swapping, while audio artifacts ($x_2$) might exhibit spectral anomalies. The shared space ($x_{12}$) captures semantic cross-modal mismatches, such as inconsistency between lip movements and speech. 
 Ideally, a robust system should integrate both perspectives for improved accuracy.
 }
 \label{fig:venn_with_face}
 \vspace{-5mm}
\end{wrapfigure}

By revisiting the existing detection works, we argue that \textit{a truly effective deepfake detection system must integrate both modality-specific and shared perspectives, ensuring that neither local artifacts nor cross-modal mismatches escape detection}. By leveraging the full multimodal space rather than treating modalities independently, we can build a stronger defense against increasingly sophisticated fakes.
As illustrated in Figure \ref{fig:venn_with_face}, we divide the traces of multimodality deepfake detection into modality-specific and modality-shared inconsistencies. \textit{The visual domain ($x_1$)} contains artifacts like blending boundaries or unrealistic motion, while the \textit{audio domain ($x_2$)} may exhibit synthetic spectral anomalies. More critically, their \textit{joint representation ($x_{12}$)} captures discrepancies that neither modality alone can reveal, such as a deepfake that synchronizes well but exhibits unnatural articulation patterns or mismatched speech emotion.

To achieve this, we propose \textbf{CAD}, a novel multimodal detection framework that unifies modality-shared semantic alignment analysis and modality-specific forensic verification. CAD operates on the \textit{principle that multimodal deepfakes inevitably leave traces in two forms}: \textbf{(1) semantic misalignments between modalities} (\eg, discrepancies between spoken words and lip movements) and (2) \textbf{modality-specific inconsistencies} (\eg, residual artifacts from facial swapping or synthetic voice generation). 
Unlike prior works that prioritize one aspect at the expense of the other, \textit{CAD introduces a dual-path architecture}: \textbf{a cross-modal alignment module to detect semantic disharmony} and \textbf{a distillation mechanism to preserve and harmonize modality-specific forensic signals}. By synergizing these complementary cues, CAD closes the detection gap left by conventional approaches, offering a unified defense against the escalating threat of multimodal synthetic media. Evaluated on the comprehensive IDForge dataset, which spans diverse forgery types, CAD achieves state-of-the-art performance (99.96\% AUC), demonstrating its robustness against evolving multimodal attacks.

Our main contributions are summarized as three-fold:
\begin{itemize}
    \item \textbf{Unified Detection of Modality-Shared and Modality-Specific Cues}: We propose CAD, a novel detection framework that integrates both modality-shared semantic alignment analysis (\eg, lip-speech mismatches) and modality-specific forensic verification (\eg, residual artifacts in face-swapped videos or synthetic audio).

    \item \textbf{Dual-Path Architecture Design}: We introduce a cross-modal alignment module to detect semantic disharmony across different modalities, and develop a cross-modal distillation mechanism that preserves forensic artifacts within individual modalities while mitigating conflicts during multimodal feature fusion.

    \item  \textbf{SOTA Performance on Multimodal Deepfake Detection}: We evaluate CAD on the IDForge dataset, which covers a wide range of deepfake manipulation techniques, achieving 99.96\% AUC, significantly outperforming existing unimodal and multimodal detection methods, validating the necessity of a holistic approach to multimodal deepfake detection.

\end{itemize}

\section{Related Works}
\subsection{Deepfake Detection}
 \noindent\textbf{Unimodal Deepfake Detection.} Most existing deepfake video detection methods focus on identifying visual artifacts in image modality~\cite{wangCVPR21rfm,yan2023ucf,shiohara2022detecting,li2020face,yan2024transcending} and video modality~\cite{Haliassos2021lips,Wang2023altfreeze,xu2023tall,zhang2024learning, yan2024generalizing}. Notably, early works, such as \cite{Haliassos2021lips}, leveraged lip-reading pre-trained models to detect inconsistencies in lip movements, while deepfake image detection initially focused on blending boundaries~\cite{li2020face}, a common artifact in face-swapping forgeries.
 Later, \cite{shiohara2022detecting, cheng2024can} further explored blending-based detection, refining the objective from cross-ID to within-ID forgery analysis~\cite{shiohara2022detecting} and examining the role of blended data in detection~\cite{cheng2024can}. Meanwhile, video-based detection evolved to balance spatial and temporal artifact learning~\cite{wang2023seeing, yan2024generalizing}, mitigating overfitting to a single modality (\eg, spatial artifacts only) and improving robustness across different forgery types.

\noindent\textbf{Multimodal Deepfake Detection.} Recent methods leverage multimodal information for deepfake detection. RealForensics~\cite{Haliassos2022Training} learns video and audio features crosswise, updating student model parameters and training classifiers. \cite{Komal2020notmf} and \cite{Cozzolino2023Audio-Visual} use both cross-entropy and contrastive learning losses to measure similarity between real samples, aiding joint feature learning. \cite{liu2024lips} focuses on the connection between audio and lip features. Building on MAE~\cite{He2022mae}, \cite{Oorloff2024AVFF} masks video and audio embeddings, using additional decoders and reconstruction loss for joint learning. \cite{Wang2024exploringdi} introduces FDMT, utilizing 3D facial information and depth-based attention to improve deepfake detection.
However, we find that most previous detection works still fail to consider how to maximize or fully leverage both modality-specific and shared forgeries for detection, leaving vulnerabilities that sophisticated attacks can exploit.

\subsection{Multimodal Representation Learning}
 With the advent of multimodal models,  CLIP~\cite{Radford2021clip} leverages contrastive learning~\cite{Chen2020cl} to align semantic information across modalities by learning cross-modal embedding representations of images and text. Building upon this approach, ImageBind~\cite{Girdhar2023imagebind} and LanguageBind~\cite{zhu2024languagebind} extend contrastive learning~\cite{Chen2020cl} to additional modalities, including vision, audio, text, depth, and more. These developments further enhance the capacity of encoders to extract features from diverse modalities, facilitating the learning of richer and more comprehensive representations.

 Vision and audio are key modalities in deepfake detection, with audio excelling in action recognition in long videos. Knowledge transfer between modalities has been explored for action recognition~\cite{Chen2021distilling}, and the BAVNet~\cite{Wu2021Binaural} localizes sound within visual scenes. AVoiD-DF~\cite{Yang2023AVoiD-DF} introduces a multimodal decoder for joint feature fusion in deepfake detection. Depth modality also aids face feature extraction. Depth maps distinguish real and fake faces~\cite{Wang2020deep}, with PRNet~\cite{Feng2018Joint3F} and contrastive depth loss for feature learning. Facial depth maps are used for anti-spoofing~\cite{Zheng2021Attention-Based}, with symmetry loss for accurate depth estimation, which is beneficial for forgery detection~\cite{Wang2024exploringdi}.
 
\section{Method}

\begin{figure*}[!t]
 \centering
 \includegraphics[width=0.9\textwidth]{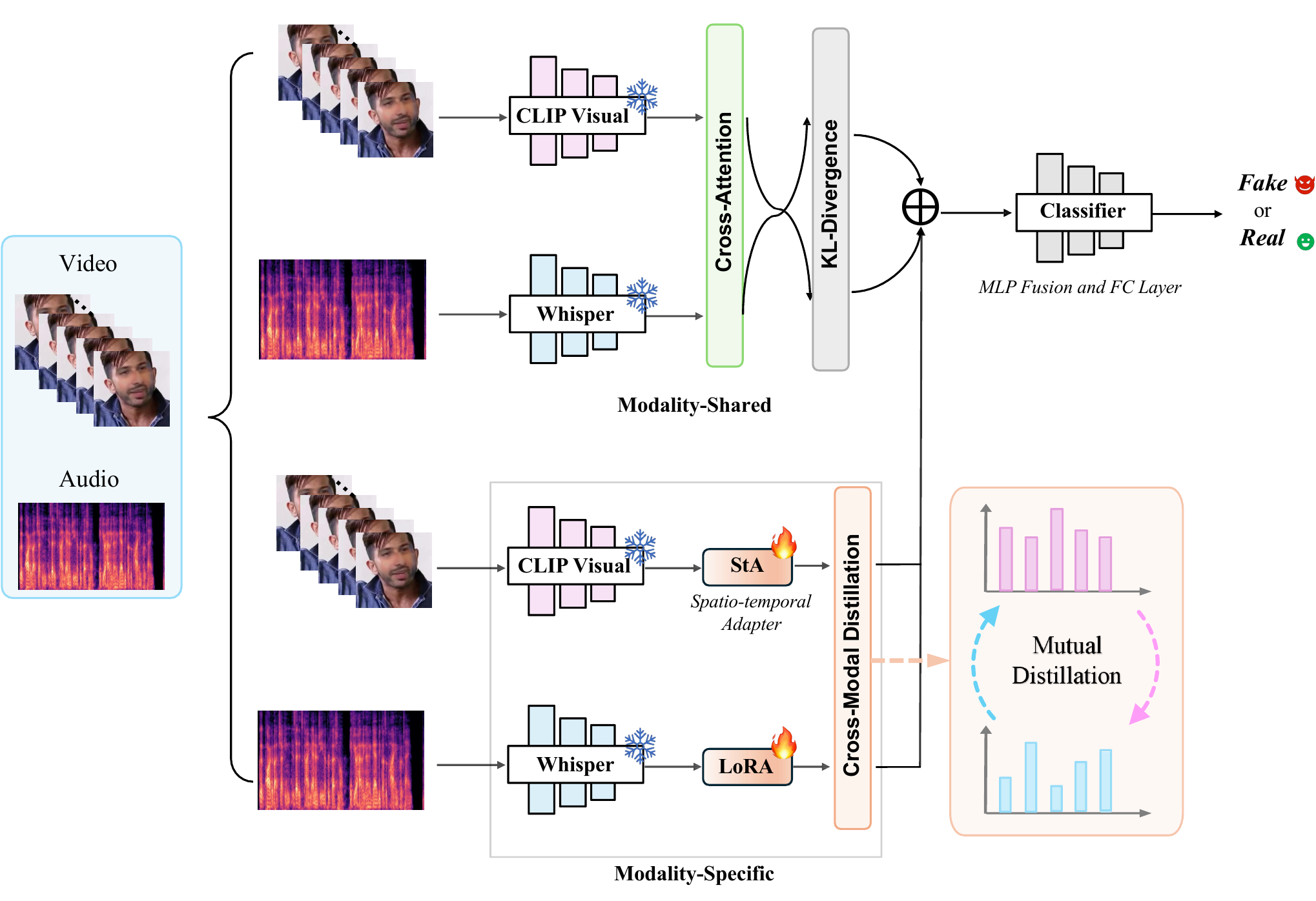}
 \caption{The overview of CAD. Our proposed CAD is designed to maximize and fully mine both modality-specific and modality-shared cues for robust deepfake detection.}
 \vspace{-7pt}
 \label{fig2:overview}
\end{figure*}

\subsection{Overview}
Multimodal deepfake detection can be addressed from \textbf{two complementary perspectives}: (1) high-level semantic cues, such as facial structure and expression consistency, and (2) low-level forensic artifacts, such as subtle pixel-level distortions. Semantic inconsistencies often manifest in facial regions involved in expression and speech (e.g., unnatural lip movements), while artifact-based traces typically appear in fine details like the nose or eye contours.

As shown in Figure~\ref{fig2:overview}, our proposed \textbf{CAD framework} moves beyond traditional approaches that focus on a single aspect of forgery detection. Instead, we unify two key components within one architecture: \textbf{modality-shared semantic alignment} and \textbf{modality-specific forensic verification}. This design captures both the high-level semantic alignment between audio and visual modalities and the low-level, modality-specific artifacts. In Section~\ref{sec:3.2}, we provide a theoretical foundation for this dual-perspective design, based on maximizing mutual information between modalities. Sections~\ref{sec:3.3} and \ref{sec:3.4} then detail the implementation of the forensic and semantic modules, each tailored to extract complementary evidence from different modalities for robust deepfake detection.

\subsection{Maximizing Multi-modality \textit{Mutual Information} in Multimodal Deepfake Detection}
\label{sec:3.2}
Multimodal deepfake detection should go beyond analyzing individual modalities in isolation. Instead, it should explicitly model potential inconsistencies between the visual and audio streams. As illustrated in Figure~\ref{fig:venn_with_face}, different deepfake techniques tend to introduce modality-specific artifacts, leading to subtle mismatches between what is seen and what is heard.
To capture these cross-modal inconsistencies, we leverage both joint and disentangled feature representations of vision ($V$) and audio ($A$), as formalized in Equation~\ref{eq1} below:
\vspace{7pt}
\begin{equation}
\label{eq1}
    H(V)=H(V\mid A)+I(V,A).
\end{equation}

Following the principles of information theory~\cite{jaynes1957information}, this formulation incorporates entropy ($H(X)$) and mutual information ($I(X, Y)$) to guide the model in learning representations that maximize shared information across modalities while preserving modality-specific signals.

Specifically, we consider the distributions of video and audio to be $x_1$ and $x_2$, respectively. We also introduce $y$, a latent space that encapsulates both $x_1$ and $x_2$, ensuring full compatibility with video and audio representations. Based on the definition of mutual information~\cite{jaynes1957information}, we can derive the following lemma below.
 
\begin{lemma}
     The mutual information between two random variables $x_1,\,x_2$ and a universal space distribution $y$ can be expressed in terms of entropy as:
\end{lemma}
\vspace{-7.6pt}
\begin{equation}
\label{eq2}
    \begin{split}
        I(x_1,&x_2,y)=H(x_1)+H(x_2)+H(y)-H(x_1,y)\\
        &\quad-H(x_2,y)-H(x_1,x_2)+H(x_1,x_2,y).
    \end{split}
\end{equation}
\vspace{2.6pt}
\begin{lemma}
    Since we define a total space distribution $y$ that integrates both modalities, ensuring seamless compatibility with their respective representations, it follows that $x_1\subset y$ and $x_2\subset y$. Consequently, the following equation is established:
\end{lemma}

\begin{equation}
\label{eq3}
    H(x_1,x_2,y)=H(x_1,y)=H(x_2,y)=H(y).
\end{equation}

Substitute into formula (1):
\vspace{8pt}
\begin{equation}
\label{eq4}
    \begin{split}
        I(x_1,x_2,y) &= H(x_1)+H(x_2)-H(x_1,x_2).
    \end{split}
\end{equation}

When we perform a single-modal task, we can consider maximizing its mutual information part, $I(x_1, y)$, which is the visual channel goal. According to the definition of mutual information, we have the following Lemma.
\begin{lemma}
     Maximizing the mutual information between the input $x_1$ and the target variable $y$ can be formulated as optimizing the information flow within a specific modality, which represents the objective for the visual channel.
\end{lemma}
\vspace{-10pt}
\begin{equation}
\label{eq5}
    I(x_1,y)=I(x_1,x_2,y)+H(x_1,y\mid x_2).
\end{equation}

Previous studies have extensively demonstrated the benefits of joint learning for multimodal feature representations. However, these approaches primarily focus on the shared joint distribution characteristics across different modalities—specifically, $I(x_1,x_2,y)$ in Equation~\ref{eq4}, while neglecting the modality-specific components $H(x_1,y\mid x_2)$ that do not align with the multimodal representation.
 In the context of audio-visual feature learning, a more detailed analysis can be conducted. Let $s$ denote a single sample, while $v$ and $a$ represent the extracted video and audio features, respectively. Furthermore, let $V$ and $A$ denote the distributions of video and audio features across all samples. Based on this formulation, we derive the following theorem.

\begin{theorem}
    For the representation learning component, when the input is encoded into an embedding vector by the encoder, the mutual information between the video and audio embedding representations of a single sample and the overall distribution is formulated as:
 \end{theorem}
\vspace{-11pt}
\begin{equation}
\label{eq6}
    \begin{split}
        I(&v,a\mid V,A) = E_{p(v)}[\,H(p\,(v\mid V))]- E_{p(s,V,A,a)}[\,H(p\,(v\mid V),\,p(v \mid a, V, A))].
    \end{split}
\end{equation}

\begin{proof}
    We can derive as follows:
    \vspace{3pt}
    \begin{equation}
    \label{eq7}
        \begin{split}
            &\quad \quad I(v,a\mid V,A) \\[4pt]
            &= \int p(V, A)\int [\,p(v,a\mid V,A) \log{\frac{p(v,a\mid V,A)}{p(v \mid V,A)p(a\mid V,A)}}] \,dv\,da\,dVdA \\ 
            &= \int p(V, A)\int [\,p(v,a\mid V,A)\log{\frac{p(v\mid a,V,A)}{p(v \mid V,A)}}] \,dv\,da\,dVdA\\ 
            &= \int p(V, A)\int [\,p(v,a\mid s,V,A)\,p(s\mid V,A)\log{\frac{p(v\mid a,V,A)}{p(v \mid V,A)}}] \,dv\,da\,dVdA \\ 
            &= \int p(s,V, A)\int [\,p(v\mid V,s)\,p(a\mid A,s)\log{\frac{p(v\mid a,V,A)}{p(v \mid V,A)}}\ \,dv\,da\,dVdA \\ 
            &=E_{p(s,V,A,a)}[\int p(v\mid V,s)\,\log{p(v\mid a,V,A)}\,dv] - E_{p(s,V,A,a,v)}[\,\log p(v\mid V,A)] \\[6pt]
            &= -E_{p(s,V,A,a)}[\,H(p(v\mid V),\,p(v\mid a,V,A))] + E_{p(v)}[\,H(p(v\mid V)]. \qedhere
        \end{split}
    \end{equation}
\end{proof}

For the first term in Equation~\ref{eq6}, the target representation must exhibit high entropy to prevent collapse and ensure adaptability to downstream tasks. The second term necessitates that the distribution of $v$ closely approximates that of $a$, thereby minimizing the distribution distance of visual and audio modalities. Notably, the second term corresponds to $I(x_1,x_2,y)$ in Equation~\ref{eq5}, which explains the effectiveness of joint learning in multimodal feature representation. In contrast, the first term aligns with traditional unimodal video authentication, as it excludes the objective of maximizing the expected joint information with audio.

 Inspired by the derivation, CAD comprehensively accounts for the influence of both components, enabling the modeling of multimodal interactions within the framework of multimodal representation learning, capturing both modality-specific characteristics and collaborative dependencies.

\subsection{Cross-Modal Alignment for Modality-Specific Forgery Learning}
\label{sec:3.3}
 In Equation~\ref{eq5}, for $I(x_1,\ x_2,\ y)$, highlights the need for learning modality-shared representations. Previous research has thoroughly validated the enhancement of feature representation through the joint learning of multi-modality. Similarly, to maximize the $I(x_1,\ x_2,\ y)$ term in Equation~\ref{eq4}, CAD effectively captures the shared information embeddings across different modalities by aligning the feature representations of audio-visual signals.

 As illustrated in the joint representation in Figure~\ref{fig2:overview}, we employ feature alignment using CLIP~\cite{Radford2021clip} and Whisper~\cite{Radford2022RobustSRwhisper} as the frozen pre-trained encoders for video and audio. After acquiring high-level semantic representations via the pre-trained model, we extract their high-level semantic features using a cross-attention mechanism, and subsequently minimize the distributional discrepancy between modalities via Kullback–Leibler (KL) divergence. Such efficient operations enable the modality-shared component to learn common semantic features across modalities, such as the correlation between lip movements and vocal signals. Let $x_{v2a}$ and $x_{a2v}$ represent the outputs after the cross-attention mechanism. And $P(x)$, $Q(x)$ represent the potential distributions of $v2a$ and $a2v$ respectively. We compute the modality-specific alignment loss as follows:


 \vspace{-10pt}
\begin{equation}
\label{eq8}
\begin{split}
    &P(x) = \text{Softmax}(x_{v2a}),\\
    &Q(x) = \text{Softmax}(x_{a2v}),\\
    &L_{\text{KL}} = \sum_{x} P(x) \log \frac{P(x)}{Q(x)}. 
\end{split}
\end{equation}

\subsection{Cross-Modal Distillation for Modality-Shared Forgery Learning}
\label{sec:3.4}
 In Equation~\ref{eq5}, for $H(x_1,y \mid x_2)$, aligns with the video modality-specific features. During the process of video feature extraction, it is crucial to consider not only the spatial representation of static frames but also to effectively capture the temporal dynamics between frames to obtain comprehensive spatiotemporal features. We draw inspiration from~\cite{yan2024generalizing} in the encoder design to optimize the spatiotemporal feature aggregation strategy. This ensures that the model can effectively integrate the local detail information of individual video frames with the global temporal patterns, thereby enhancing the overall performance of video understanding tasks. We employ the audio pre-trained frozen model with Low-Rank Adaptation (LoRA)~\cite{hu2022lowrank}, allowing efficient fine-tuning through a trainable low-rank decomposition matrix. This approach enables effective adaptation to task-specific requirements with minimal computational overhead. 

 As illustrated in the specific representation in Figure~\ref{fig2:overview}, to leverage the audio modality to enhance video representation learning and improve cross-modal consistency, we applies distillation between the two unimodal channels. As the distillation loss, we introduce the SimSiam~\cite{chen2021exploring} loss, which offers two key advantages. First, SimSiam mitigates the introduction of erroneous negative samples by directly comparing representations, thereby addressing potential distributional discrepancies between different modalities. Second, it prevents feature collapse by ensuring that representations from different modalities do not converge into identical vectors but instead retain modality-specific information, preserving the unique characteristics of each unimodal representation. Denote $x_v^u$ and $x_a^u$ as the unimodal embeddings for vision and audio, respectively, and $z_v^u$ and $z_a^u$ as the SimSiam projections for vision and audio. $\odot$ represents element-wise multiplication and \textit{Norm} represents L2 normalization. We compute the modality-shared cross-modal distillation loss as:
\vspace{3pt}
\begin{equation}
\label{eq9}
\begin{split}
    & Dist_1 = -Norm(x_v^u)\odot Norm(z_a^u), \\
    & Dist_2 = -Norm(x_a^u)\odot Norm(z_v^u), \\
    & L_{\text{KD}} = (Dist_1+Dist_2)/2.
\end{split}
\end{equation}

\section{Experiments}
\label{sec:4}
\subsection{Implementation Settings}
\label{sec:4.1}
 \noindent\textbf{Dataset.} To comprehensively assess the performance of CAD from multiple perspectives, we conducted experiments on a diverse range of audio-visual and video-only deepfake datasets. Specifically, we utilized the following datasets: (1) \textbf{FakeAVCeleb~\cite{Khalid2021FakeAVCeleb}.} This dataset comprises both video and audio deepfakes, featuring precisely synchronized lip movements and fine-grained annotations. (2) \textbf{IDForge-v2~\cite{xu2024idforge}}. IDForge-v2 improves upon IDForge-v1 by adding compression and super-resolution to better simulate real-world conditions. Each video retains 16 frames in four evenly spaced groups, along with corresponding audio and text. (3) \textbf{FaceShifter~\cite{Rössler2019FF+} and Celeb-DF~\cite{Li2020celebdf}.} FaceShifter and Celeb-DF consist exclusively of visual deepfakes and enhance various forgery approaches to generate synthetic videos.

\begin{wraptable}{r}{0.48\textwidth}
\renewcommand{\arraystretch}{1.1}
	\centering
    \vspace{-5pt}
    \captionof{table}{\textbf{Intra-manipulation evaluation} on \textbf{FakeAVCeleb~\cite{Khalid2021FakeAVCeleb}}.  Following~\cite{Yang2023AVoiD-DF, Oorloff2024AVFF}, select 70\% as the training set and the remaining 30\% as the test set. We report accuracy (Acc) and Area Under the Curve (AUC) scores (\%).}
    \label{tab:ablation fakeav30_70}
    \resizebox{0.99\linewidth}{!}{
     \begin{tabular}{c|c|cc}
    \toprule
    \multicolumn{1}{c|}{\multirow{1}*{Method}} & \multicolumn{1}{c|}{\multirow{1}*{Modality}} & \multicolumn{1}{c}{Acc} & \multicolumn{1}{c}{AUC} \\
    \midrule
    \text{MesoNet~\cite{Afchar2018MesoNet}} & \textcolor{lightred}{V} & 57.3 & 60.9 \\
    \text{Capsule}~\cite{Nguyen2019Capsule} & \textcolor{lightred}{V} & 68.8 & 70.9 \\
    \text{Head Pose}~\cite{Yang2019exposing} & \textcolor{lightred}{V} & 45.6 & 49.2 \\
    \text{VA-MLP}~\cite{Matern2019Exploiting} & \textcolor{lightred}{V} & 65.0 & 67.1 \\
    \text{Xception}~\cite{Chollet2017xception} & \textcolor{lightred}{V} & 67.9 & 70.5 \\
    \text{LipForensics}~\cite{Haliassos2021lips} & \textcolor{lightred}{V} & 80.2 & 82.4 \\
    \text{DeFakeHop}~\cite{Chen2021defakehop} & \textcolor{lightred}{V} & 68.3 & 71.6 \\
    \text{CViT}~\cite{Wodajo2021DeepfakeVD} & \textcolor{lightred}{V} & 70.5 & 72.1 \\
    \text{Multiple-Attention}~\cite{Zhao2021MultiattentionalDD} & \textcolor{lightred}{V} & 77.6 & 79.3 \\
    \text{SLADD}~\cite{Chen2022self-su} & \textcolor{lightred}{V} & 70.5 & 72.1 \\
    \midrule
    \text{AVN-J}~\cite{Qian2021audio-visual} & \textcolor{lightgreen}{A-V} & 73.2&77.6 \\
    \text{MDS}~\cite{Komal2020notmf} & \textcolor{lightgreen}{A-V} & 82.8 &86.5 \\
    \text{Emotion Don't Lie}~\cite{Mittal2020EmotionsDL} & \textcolor{lightgreen}{A-V} & 78.1&79.8 \\
    \text{AVFakeNet}~\cite{Ilyas2023AVFakeNetAU} & \textcolor{lightgreen}{A-V} & 78.4 & 83.4 \\
    \text{VFD}~\cite{Cheng2022VoiceFaceHT} & \textcolor{lightgreen}{A-V} & 81.5 & 86.1 \\
    \text{BA-TFD}~\cite{Cai2022Content} & \textcolor{lightgreen}{A-V} & 80.8 & 84.9 \\
    \text{AVoiD-DF}~\cite{Yang2023AVoiD-DF} & \textcolor{lightgreen}{A-V} & 83.7 & 89.2 \\
    \text{AVFF}~\cite{Oorloff2024AVFF} & \textcolor{lightgreen}{A-V} & 98.6 & 99.1 \\
    \midrule
    \textbf{CAD (Ours)} & \textcolor{lightgreen}{A-V} & \textbf{99.0} & \textbf{99.6} \\
    \bottomrule
	\end{tabular}}
\vspace{-35pt}
\end{wraptable}
 
 \noindent\textbf{Training details.} During training, 16 frames per sample are used without truncation to align with Whisper’s 30-second audio feature extraction, preserving temporal continuity. For datasets with extensive backgrounds, facial landmarks are detected and cropped beforehand. We trained on 8$\times$NVIDIA-H100 for 4 hours.

\begin{wraptable}{r}{0.48\textwidth}
\renewcommand{\arraystretch}{1.2}
	\centering
	\caption{\textbf{Intra-manipulation evaluation on IDForge~\cite{xu2024idforge}.} We conducted experiments based on the training and test sets divided by IDForge-v2. IDForge contains 11 different forgery subsets and we conduct multiple tests and compute the average for robustness. We report Area Under the Curve (AUC) scores (\%) and Average Precision (AP) scores (\%) .}
    \label{tab:ablation idforge}
    \vspace{-.5em}
	 \resizebox{1.0\linewidth}{!}{
    \setlength\tabcolsep{12pt}
     \begin{tabular}{c|c|cc}
    \toprule
    \multicolumn{1}{c|}{\multirow{1}*{Method}} & \multicolumn{1}{c|}{\multirow{1}*{Modality}} & \multicolumn{1}{c}{AUC} & \multicolumn{1}{c}{AP} \\
    \midrule 
    \text{MesoI4~\cite{Afchar2018MesoNet}}&\textcolor{lightred}{V}&74.72&38.39\\
    \text{I3D~\cite{Carreira2017QuoVA}}&\textcolor{lightred}{V}&77.74&45.56\\
    \text{CADDM~\cite{Dong2022ImplicitIL}}&\textcolor{lightred}{V}&80.35&59.92\\
    \text{UFD~\cite{Ojha2023Towards}}&\textcolor{lightred}{V}&89.12&70.00\\
    \text{RealForensics~\cite{Haliassos2022Training}}&\textcolor{lightred}{V}&93.21&88.18\\
    \midrule
    \text{CDCN~\cite{Khalid2021EvaluationOA}}&\textcolor{lightgreen}{A-V}&87.43&71.17\\
    \text{VFD~\cite{Cheng2022VoiceFaceHT}}&\textcolor{lightgreen}{A-V}&90.70&-\\
    \text{IDForge~\cite{xu2024idforge}}&\textcolor{lightgreen}{A-V}&98.12&86.67\\
    \midrule
    \textbf{CAD (Ours)} &\textcolor{lightgreen}{A-V}&\textbf{99.96}&\textbf{99.63}\\
    \bottomrule
	\end{tabular}}
    \vspace{-10pt}
\end{wraptable}

\noindent\textbf{Model details.} For the pre-trained models, we use CLIP ViT-Base-16~\cite{Radford2021clip} and Whisper-Small~\cite{Radford2022RobustSRwhisper} to balance prior knowledge and computational efficiency. Videos are preprocessed via the CLIP processor, resizing to 224$\times$224 and normalizing RGB channels. The LoRA~\cite{hu2022lora} decomposition in the audio encoder defaults to $r=8$ and $\alpha=16$ unless specified.

\subsection{Comparison with Existing Methods}
  We evaluate our model's performance against state-of-the-art algorithms using multiple criteria, including average accuracy (Acc), Average Precision (AP), and Area Under the Curve (AUC), across diverse datasets. For audio-visual algorithms, a video is classified as fake if either modality, or both, is manipulated. Unimodal models consider a video fake only when the visual modality is forged.

 \noindent\textbf{Intra-manipulation evaluations.} We follow the methodology outlined in AVoiD-DF~\cite{Yang2023AVoiD-DF}, randomly selecting 70\% of FakeAVCeleb~\cite{Khalid2021FakeAVCeleb} as the training set for model training, while the remaining 30\% serves as the evaluation set. As shown in Table~\ref{tab:ablation fakeav30_70}, our method demonstrates significant improvements over two audio-visual deepfake detection methods, AVoiD-DF~\cite{Yang2023AVoiD-DF} and AVFF~\cite{Oorloff2024AVFF}. Table~\ref{tab:ablation idforge} further presents a comparative analysis between CAD and multiple open-source state-of-the-art forgery detection methods on the IDForge dataset~\cite{xu2024idforge}, encompassing both unimodal and multimodal approaches. 

\begin{table}[!t]
\renewcommand{\arraystretch}{1.2}
	\centering
    \caption{\textbf{Cross-manipulation evaluation} on \textbf{FakeAVCeleb~\cite{Khalid2021FakeAVCeleb}}. The performance is evaluated using a leave-one-category-out approach, where one category is reserved for testing while the remaining categories are used for training. We report Average Precision (AP) scores (\%) and Area Under the Curve (AUC) scores (\%). The average performance category is provided in AVG-FV.}
    \label{tab:ablation-fakeav}
    \vspace{7pt}
    \resizebox{1.0\linewidth}{!}{
    \setlength\tabcolsep{8pt}
     \begin{tabular}{c|cc|cccccccccccc}
    \toprule
    \multicolumn{1}{c|}{\multirow{2}*{Method}} & \multicolumn{2}{c|}{\multirow{2}*{Modality}} & \multicolumn{2}{c@{\hspace{3mm}}}{RVFA} & \multicolumn{2}{c}{FVRA-WL} & \multicolumn{2}{c}{FVFA-FS} & \multicolumn{2}{c}{FVFA-GAN} & \multicolumn{2}{c}{FVFA-WL} & \multicolumn{2}{c}{AVG-FV} \\
    \cline{4-15} 
      &  &  & AP &AUC & AP& AUC & AP& AUC & AP& AUC & AP& AUC & AP& AUC \\
    \midrule
    Xception~\cite{Rössler2019FF+} & \multicolumn{2}{c|}{\textcolor{lightred}{V}} & - & - & 88.2& 88.3&92.3&93.5&67.6&68.5&91.0&91.0&84.8&85.3\\
    LipForensics~\cite{Haliassos2021lips} & \multicolumn{2}{c|}{\textcolor{lightred}{V}} &-&-&97.8&97.7&99.9&99.9&61.5&68.1&98.6&98.7&89.4&91.1 \\
    FTCN~\cite{Zheng2021ExploringTC} & \multicolumn{2}{c|}{\textcolor{lightred}{V}} &-&-&96.2&97.4&100&100&77.4&78.3&95.6&96.5&92.3&93.1\\
    RealForensics~\cite{Haliassos2022Training} & \multicolumn{2}{c|}{\textcolor{lightred}{V}} &-&-&88.8&93.0&99.3&99.1&99.8&99.8&93.4&96.7&95.3&97.1\\
    \midrule
    AV-DFD~\cite{Zhou2021joint} & \multicolumn{2}{c|}{\textcolor{lightgreen}{A-V}} & 74.9&73.3&97.0&97.4&99.6&99.7&58.4&55.4&100&100&88.8&88.1\\
    AVAD(LRS2)~\cite{Feng2023self} & \multicolumn{2}{c|}{\textcolor{lightgreen}{A-V}} &62.4&71.6&93.6&93.7&95.3&95.8&94.1&94.3&93.8&94.1&94.2&94.5\\
    AVAD(LRS3)~\cite{Feng2023self} & \multicolumn{2}{c|}{\textcolor{lightgreen}{A-V}} &70.7&80.5&91.1&93.0&91.0&92.3&91.6&92.7&91.4&93.1&91.3&92.8\\
    \text{AVFF~\cite{Oorloff2024AVFF}} &\multicolumn{2}{c|}{\textcolor{lightgreen}{A-V}} &93.3&92.4&94.8&98.2&100&100&99.9&100&99.4&99.8&98.5&99.5 \\
    \midrule
    \textbf{CAD (Ours)} & \multicolumn{2}{c|}{\textcolor{lightgreen}{A-V}}& \textbf{99.9}&\textbf{99.9}&\textbf{100} &\textbf{100} &\textbf{100} &\textbf{100} &\textbf{100} &\textbf{100} &\textbf{100} &\textbf{100} &\textbf{100} &\textbf{100} \\
    \bottomrule
	\end{tabular}}
    \vspace{-8pt}
\end{table}

For all models except VFD, additional classification heads are incorporated during experiments to facilitate classification. The results indicate that CAD outperforms RealForensics significantly, with a 6.75\% increase in AUC and an 11.45\% improvement in AP. Additionally, compared to the R-MFDN network proposed in IDForge, CAD demonstrates notable superiority, achieving a 12.96\% increase in AP.

 \noindent\textbf{Cross-manipulation evaluations.} The cross-manipulation evaluation examines model performance on a dataset with an unknown distribution in Table~\ref{tab:ablation-fakeav}. emphasizing the necessity for deepfake detection algorithms to exhibit robust generalization capabilities against previously unseen forgery techniques to ensure adaptability across diverse scenarios. To systematically evaluate this aspect, we categorize subsets of FakeAVCeleb~\cite{Khalid2021FakeAVCeleb} based on different forgery techniques, including RVFA, 

\renewcommand{\arraystretch}{1.2}
\begin{wraptable}{r}{0.6\textwidth}
    \vspace{0pt}
	\centering
	\caption{Ablation experiments of the proposed \textbf{cross-modal alignment} with training on IDForge-v2, we report Area Under the Curve (AUC) scores (\%) and Average Precision (AP) scores(\%) and test on FaceForensics++ and Celeb-DF.}
    \label{tab:self-module alignment ablation cross manipulation}
    \vspace{-.5em}
	\resizebox{1.0\linewidth}{!}{
    \setlength\tabcolsep{6pt}
     \begin{tabular}{c|cc|cc|cc}
    \toprule
    \multicolumn{1}{c|}{\multirow{2}*{Method}} & \multicolumn{2}{c|}{\multirow{1}*{FaceShifter}} & \multicolumn{2}{c|}{\multirow{1}*{Celeb-DF}} & \multicolumn{2}{c}{\multirow{1}*{Avg}} \\
    & AUC & AP & AUC & AP & AUC & AP\\
    \midrule
    \text{\textbf{only} video} & 100 & 100 & 74.9 & 80.5 & 87.5 & 90.3 \\
    \text{$w/o$ alignment} &98.6 & 98.8 & 93.5 & 96 & 96.1 & 97.4 \\
    \text{$w/o$ cross attention} & 95.7 & 95.9 & 85.1 & 90.1 & 90.4 & 93\\
    \text{$w/o$ frozen model} & 99.9 & 99.9 & 90.8 & 95.0 & 95.4 & 97.5\\
    \midrule
    \textbf{CAD (Ours)} &\textbf{99.6}& \textbf{99.6} & \textbf{94.2} & \textbf{96.7} & \textbf{96.9} & \textbf{98.2}\\
    \bottomrule
	\end{tabular}}
\end{wraptable}
 
To assess the generalization ability of our model, we conduct cross-category evaluation on four distinct forgery types: FVRA-WL, FVFA-FS, FVFA-GAN, and FVFA-WL, each representing a unique data generation paradigm. In each experiment, one forgery category is excluded during training and used exclusively for testing, while the model is trained on the remaining categories.
As shown in Table~\ref{tab:ablation-fakeav}, CAD exhibits strong generalization capabilities, consistently outperforming baseline methods when facing unknown forgery types. This robustness highlights the effectiveness of our joint modeling of semantic alignment and modality-specific forensic cues.
Furthermore, results across Table~\ref{tab:ablation fakeav30_70} and Table~\ref{tab:ablation idforge} reinforce a clear trend: multimodal deepfake detection methods consistently surpass unimodal approaches in both accuracy and robustness. 

\renewcommand{\arraystretch}{1.3}
\begin{wraptable}{r}{0.6\textwidth}
    \vspace{-11pt}
	\centering
	\caption{Ablation experiments of the proposed \textbf{cross-modal distillation} with training on IDForge-v2, we report Area Under the Curve (AUC) scores (\%) and Average Precision (AP) scores(\%) and test on FaceForensics++ and Celeb-DF.}
    \label{tab:self-module distil ablation cross manipulation}
    \vspace{-.5em}
	\resizebox{1.0\linewidth}{!}{
    \setlength\tabcolsep{6pt}
     \begin{tabular}{c|cc|cc|cc}
    \toprule
    \multicolumn{1}{c|}{\multirow{2}*{Method}} & \multicolumn{2}{c|}{\multirow{1}*{FaceShifter}} & \multicolumn{2}{c|}{\multirow{1}*{Celeb-DF}} & \multicolumn{2}{c}{\multirow{1}*{Avg}} \\
    & AUC & AP & AUC & AP & AUC & AP\\
    \midrule
    \text{\textbf{only} video} & 100 & 100 & 74.9 & 80.5 & 87.5 & 90.3 \\
    \text{$w/o$ distillation} &97.8& 97.5 & 87.9& 91.6 & 92.9 & 94.6\\
    \text{Substitute SimSiam with KL} & 99.2 & 99.6 & 93.2 & 95.8 & 96.2 & 97.7\\
    \midrule
    \textbf{CAD (Ours)} &\textbf{99.6}& \textbf{99.6} & \textbf{94.2} & \textbf{96.7} & \textbf{96.9} & \textbf{98.2}\\
    \bottomrule
	\end{tabular}}
    \vspace{-15pt}
\end{wraptable}

\subsection{Ablation Study}
To investigate the rationality of the CAD architecture for video detection, we conducted a series of self-module ablation experiments. As presented in Table~\ref{tab:self-module alignment ablation cross manipulation} and Table~\ref{tab:self-module distil ablation cross manipulation}, the model was trained on IdForge~\cite{xu2024idforge} and evaluated on FaceShifter of FaceForensics++~\cite{Rössler2019FF+} and CelebDF~\cite{Li2020celebdf}, which both lack audio channels. The results in Table~\ref{tab:self-module alignment ablation cross manipulation} and Table~\ref{tab:self-module distil ablation cross manipulation} indicate that the performance of the model utilizing only modality-specific features is significantly inferior to that with dual modality inputs, highlighting the critical role of multimodal information complementarity in enhancing single-modality learning. 
The absence of cross-modal distillation results in a decrease of 4\% and 3.6\% in average AUC and AP, respectively, indicating a clear drop in detection performance. This highlights the effectiveness of cross-modal distillation in alleviating distributional discrepancies between modalities and enhancing the discriminability of the learned feature representations.
In addition, substituting the distillation strategy with a KL divergence loss leads to a further decline in performance, suggesting that KL divergence may be suboptimal for capturing complex multimodal relationships in this context.

\begin{figure}[t!]
 \centering
 \includegraphics[width=\linewidth]{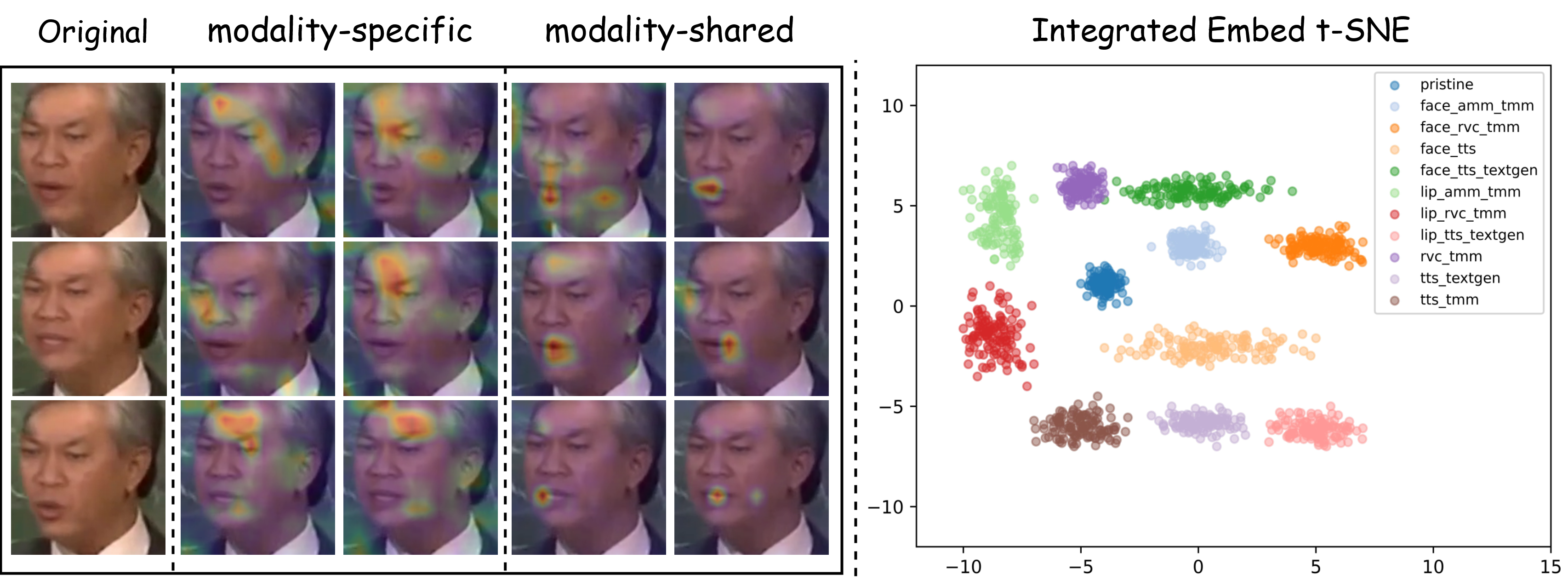}
 \vspace{-5pt}
 \caption{Visual illustrations of our method. \textit{\textbf{Left}: The visualization results of modality-specific learning and modality-shared learning by CAM~\cite{Zhou2015LearningDF}.} \textbf{Origin} denotes the original video input. Modality-specific shows the attention distribution within the vision unimodal encoder, where attention focuses mainly on visual artifacts. Instead, Modality-shared illustrates the attention distribution when both modalities are aligned, with attention primarily on the lips and surrounding musculature.
 \textit{\textbf{Right}: t-SNE visualization on integrated embeddings.}}
 \label{fig:visualization}
 \vspace{-3mm}
\end{figure}

\subsection{More Visualization and Analysis}
 We further conducted a visualization experiment and performed more analyses. As depicted in Figure~\ref{fig:visualization}, we extracted the attention weights from the model and overlaid them onto the input image to examine the model's attention distribution across different input regions. The term \textbf{modality-specific} refers to the encoder weights corresponding to the video unimodality in the CAD. While \textbf{modality-shared} denotes the attention weight distribution during joint learning. Observations with training on IDForge-v2, specific vision modal focuses on pixel-level forgery traces around the eyes and nose, and these forgeries are typically the result of manipulations in the video source. The joint learning component places greater emphasis on the lips and the surrounding muscle tremors, and these tremors highlight the regions where general features capture shared high-level semantics across modalities. 

 The attention distribution in Figure~\ref{fig:visualization} demonstrates the necessity of modality-shared semantic alignment analysis and modality-specific forensic verification, consistent with the conclusion of Theorem 1.
 In Figure~\ref{fig:visualization}, we present t-SNE visualizations of various forgery subsets from the IDForge test set. The embeddings represent the final integrated vector from CAD, encompassing both the specific and shared components. It is evident that each subset forms a distinct cluster, suggesting that CAD effectively captures features corresponding to different forgery methods.


\section{Conclusion}
In this paper, we introduce CAD, a novel method that leverages mutual information maximization to enhance multimodal alignment and trans-membrane state distillation. Our approach integrates two key components: modality-shared semantic alignment analysis and modality-specific forensic verification, enabling the extraction and complementation of information across different states. This provides a fresh perspective for the design of face authentication models. CAD is the first framework to jointly perform both coupled and decoupled feature learning in the domain of audio-visual authentication. It incorporates KL divergence and distillation loss, achieving state-of-the-art performance on the IDForge benchmark. Our findings not only demonstrate the effectiveness of CAD, but also offer a promising solution to mitigate the escalating risks associated with face forgery.

\label{sec:limit}
\textbf{Limitations and Future Work:} While CAD effectively integrates modality-shared semantics and modality-specific forensics, its current fusion and alignment mechanisms are relatively constrained in their modeling capacity. In future work, we plan to explore much larger models and integrate auto-regressive LLMs to enable more expressive and context-aware modeling of semantic consistency and cross-modal relationships. This could potentially further enhance the framework’s ability to reason over complex temporal cues and subtle modality inconsistencies in challenging forgery scenarios.

\bibliographystyle{plain}
\bibliography{ref}

\end{document}